\documentclass[lettersize,journal]{IEEEtran}
\usepackage{amsmath,amsfonts}
\usepackage{algorithmic}
\usepackage{algorithm}
\usepackage{array}
\usepackage[caption=false,font=normalsize,labelfont=sf,textfont=sf]{subfig}
\usepackage{textcomp}
\usepackage{stfloats}
\usepackage{url}
\usepackage{bm}
\usepackage{verbatim}
\usepackage{booktabs}
\usepackage{multirow}
\usepackage{graphicx}
\usepackage{cite}

\begin{document}

\title{Dynamic Contrastive Learning for Hierarchical Retrieval: A Case Study of Distance-Aware Cross-View Geo-Localization}

\author{Suofei Zhang, Xinxin Wang, Xiaofu Wu, Quan Zhou and Haifeng Hu
\thanks{This paper was produced by the IEEE Publication Technology Group. They are in Piscataway, NJ.}
\thanks{Suofei Zhang is with the School of Internet of Things, Nanjing University of Posts and Telecommunications, Nanjing 210003, China (e-mails: zhangsuofei@njupt.edu.cn).}
\thanks{Xinxin Wang, Xiaofu Wu, Quan Zhou and Haifeng Hu are with the National Engineering Research Center of Communications and Networking, Nanjing University of Posts and Telecommunications, Nanjing 210003, China (e-mails: 1022010111@njupt.edu.cn; xfuwu@ieee.org; quan.zhou@njupt.edu.cn; huhf@njupt.edu.cn;).}
}

\markboth{Journal of \LaTeX\ Class Files,~Vol.~14, No.~8, August~2021}%
{Shell \MakeLowercase{\textit{et al.}}: A Sample Article Using IEEEtran.cls for IEEE Journals}

\IEEEpubid{0000--0000/00\$00.00~\copyright~2021 IEEE}

\maketitle 

\begin{abstract}
  Existing deep learning-based cross-view geo-localization methods primarily focus on improving the accuracy of cross-domain image matching, rather than enabling models to comprehensively capture contextual information around the target and minimize the cost of localization errors.
  To support systematic research into this Distance-Aware Cross-View Geo-Localization (DACVGL) problem, we construct Distance-Aware Campus (DA-Campus), the first benchmark that pairs multi-view imagery with precise distance annotations across three spatial resolutions.
  Based on DA-Campus, we formulate DACVGL as a hierarchical retrieval problem across different domains.
  Our study further reveals that, due to the inherent complexity of spatial relationships among buildings, this problem can only be addressed via a contrastive learning paradigm, rather than conventional metric learning.
  To tackle this challenge, we propose Dynamic Contrastive Learning (DyCL), a novel framework that progressively aligns feature representations according to hierarchical spatial margins.
  Extensive experiments demonstrate that DyCL is highly complementary to existing multi-scale metric learning methods and yields substantial improvements in both hierarchical retrieval performance and overall cross-view geo-localization accuracy.
  Our code and benchmark are publicly available at https://github.com/anocodetest1/DyCL.
\end{abstract}

\section{Introduction} 
\IEEEPARstart{T}{his} paper explores the task of representation learning for Cross-View Geo-Localization (CVGL) and further extends this paradigm to the Hierarchical Retrieval (HR) scenario, thereby advancing a more pertinent and robust CVGL framework.
Given a query image captured from one viewpoint (such as the ground), the standard cross-view geo-localization problem~\cite{7299135,8099699,7850985,3996808,9607674} is defined as identifying the corresponding images of the same geographic location in a reference database, where the reference images are acquired from a different viewpoint (such as satellite).
From a theoretical perspective, cross-view geo-localization can be formulated as a cross-domain image retrieval task~\cite{7780941,4270197,8607049,1631298}.
Within the deep learning literature, metric learning~\cite{8578650,9449988,10484470} has emerged as a widely adopted approach to address this challenge, enabling neural networks to learn view-invariant feature representations. 
These models construct an embedding space where visual features from disparate domains are aligned, allowing for quantitative similarity comparisons.

In practical applications, image retrieval methods typically generate a ranked list of candidate images in descending order of similarity to the query.
The performance of such methods is evaluated based on the principle that relevant images—those depicting the same real-world location—should appear as early as possible in the ranked list.
Basically, this evaluation relies on semantic labels, which assign each image to a predefined category or identity (e.g., a specific building or landmark).
Established benchmarks, such as CVUSA~\cite{workman2015localize}, CVACT~\cite{8954224}, and University-1652~\cite{zheng2020university,zheng2023uavm}, predominantly employ this semantic label-based evaluation protocol and provide the corresponding annotation information.

In this paper, we revisit the limitations of this paradigm.
Semantic label-based methods can only indicate whether a reference image is an exact match to the query, but fail to capture the geographic relationship between the query and candidate images.
We argue that such relationships are also important in realistic cross-view geo-localization scenarios.
For example, as shown in Fig. 1, given a query image of a specific building (marked as 1 in green), \textit{exact-match} localization successfully places the true matches at the top of the retrieval list.
However, when no other exact matches exist, this approach often assigns high ranks to distant, unrelated buildings (marked in red), simply due to the absence of better alternatives.
In contrast, a \textit{distance-aware} approach prioritizes not only the exact matches but also buildings that are geographically close to the query (marked in yellow), ensuring that images physically proximate to the target are ranked ahead of distant, irrelevant ones. 
This ranking strategy offers clear advantages: even when the model makes mistake, the top-ranked images are more likely to contain potential clues or contextual information about the target location, thus resulting in more pertinent geo-localization.
Building on this motivation, we introduce the concept of \textit{Distance-Aware Cross-View Geo-Localization} (DACVGL), which explicitly incorporates spatial proximity into the retrieval process to generate rankings that are both geographically meaningful and semantically robust.
\begin{figure*}[!t]
  \centering
  \includegraphics[width=0.95\textwidth]{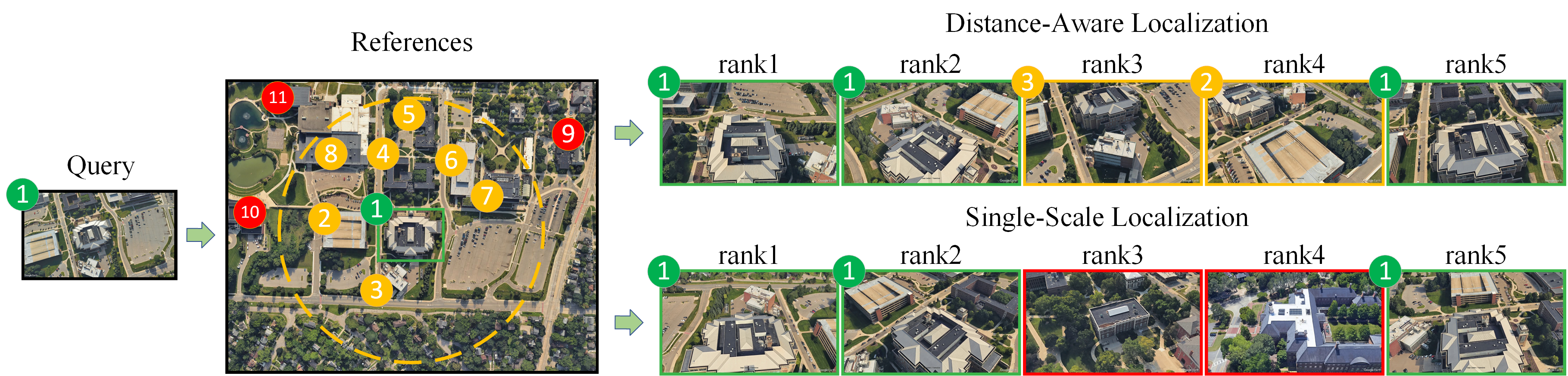}
  \caption{Illustration of a standard cross-view geo-localization retrieval process. 
  Given a query image of a specific building marked in green, the system ranks candidates from a reference set that includes both the target and surrounding buildings. 
  Two types of retrieval results are compared: \textit{exact-match localization} and \textit{distance-aware localization}. 
  The term \textit{exact-match localization} refers to the conventional geo-localization paradigm, where only images that precisely match the semantic label of the query are considered relevant.
  To the contrary, the proposed \textit{distance-aware localization} explicitly incorporates spatial proximity into the ranking process.
  In the distance-aware ranking, rank 3 and rank 4 candidates (marked in yellow) are mismatched samples as well.
  However, it is worth noting that these samples contain partial views or contextual cues of the true target in the background, thus providing potential information for more pertinent geo-localization.}
  \label{fig_1}
  \end{figure*}

To the best of our knowledge, existing cross-view geo-localization methods~\cite{huangCVCities2024,10378384,zhu2023simpleeffectivegeneralnew,3548102} focus exclusively on image matching at a specific semantic scale.
They neither provide a theoretical framework to model the geographic relationships in retrieval rankings, nor offer a dedicated benchmark to quantitatively assess how well algorithms capture spatial proximity.
On the other hand, the concept “dynamic range” has been explored in the context of HR~\cite{9578883,10.1007/978-3-031-19781-9_15}.
For instance, by Dynamic Metric Learning (DyML), two images of “Elk” from difference instances can be treated as dissimilar at the fine scale and similar at the coarse scale of “Animals”, simultaneously in a unified feature space.
Inspired by this, we introduce dynamic metric learning into cross-view geo-localization and formally define the problem of DACVGL.

However, note that DACVGL still differs essentially from standard DyML. 
First, as a cross-view matching task, it requires feature alignment between samples from two distinct domains. 
Second, conventional DyML in HR assumes explicit semantic labels and a static tree-structured correspondence across all scales. 
In contrast, in DACVGL, explicit semantic labels are only provided at the smallest scale.
Higher-level relationships among samples can only be represented by ranking based on geographic distance, rather than by predefined semantic hierarchies.
  
To address these challenges, we propose the \textit{Dynamic Contrastive Learning} (DyCL) approach, enabling the model to learn cross-view hierarchical retrieval without explicit semantic supervision. 
In addition, to facilitate systematic evaluation of how different localization methods understand spatial relationships, we construct the Distance-Aware Campus (DA-Campus) dataset. 
Analogous to University-1652~\cite{zheng2020university,zheng2023uavm}, DA-Campus features a carefully designed spatial distribution of buildings, with multi-view images collected from both drone and satellite perspectives.
Furthermore, every image in DA-Campus is geo-referenced with precise GPS tags, enabling spatial relationships between buildings to be quantified by distance and thus supporting hierarchical evaluation at multiple scales.

To sum up, the main contributions of this paper are as follows:
\begin{itemize}
  \item We formally propose the DACVGL problem and construct the DA-Campus dataset, which enables hierarchical evaluation of spatial relationships in cross-view retrieval tasks. The dataset is benchmarked with state-of-the-art retrieval and geo-localization methods.

  \item We introduce the DyCL approach to address DACVGL, enabling more effective modeling of geographic relationships during retrieval.
  
  \item As a minor contribution, we further develop a multi-scale re-ranking method, which further improves retrieval performance across different scales of spatial granularity.
  
  \item Extensive experiments demonstrate the efficacy of our approach and the value of the DA-Campus benchmark for advancing research in cross-view geo-localization.

  \end{itemize}

\section{Related Work}
\subsection{Cross-View Geo-Localization}
Recent progress in cross-view geo-localization has largely been driven by deep learning models with increasingly sophisticated network architectures and optimization strategies.
The introduction of NetVLAD~\cite{7937898} into siamese frameworks~\cite{8578856} enabled the extraction of descriptors robust to large viewpoint variations. 
Subsequent research on network architectures introduced several key enhancements, including orientation encoding~\cite{8954224}, spatial layout modeling~\cite{shi2019optimalfeaturetransportcrossview}, domain alignment and spatial attention~\cite{NEURIPS2019_ba2f0015}, as well as dynamic similarity matching modules~\cite{9157033}, all of which have contributed to improving cross-view feature representation.
From the perspective of model training, specialized loss functions such as ranking losses~\cite{10.1007/978-3-319-46448-0_30} and instance loss~\cite{zheng2020university} are exploited to achieve competitive retrieval performance.
Also, Sample4Geo~\cite{10378384} explored hard negative sampling strategies, demonstrating that effective negative mining significantly improves model discrimination in cross-view matching.

More recently, research in cross-view geo-localization has focused on the design and adaptation of backbone architectures, either by developing specialized backbone networks tailored for this task or by leveraging large-scale vision foundation models for higher performance.
\cite{zhu2023simpleeffectivegeneralnew} proposed a streamlined and generalizable backbone architecture, achieving strong performance across various geo-localization tasks.
The CV-Cities dataset~\cite{huangCVCities2024} introduced a large-scale benchmark covering major world cities, supporting research on cross-view geo-localization in complex urban environments. 
Leveraging this dataset, they demonstrated that large-scale vision foundation models can achieve strong performance for cross-view geo-localization.
In addition, 
Despite these advances, current methods primarily focus on whether the retrieval result is correct, with little attention paid to the error cost or spatial implications of incorrect predictions.
\begin{figure}[!t]
  \centering
  \includegraphics[width=0.45\textwidth]{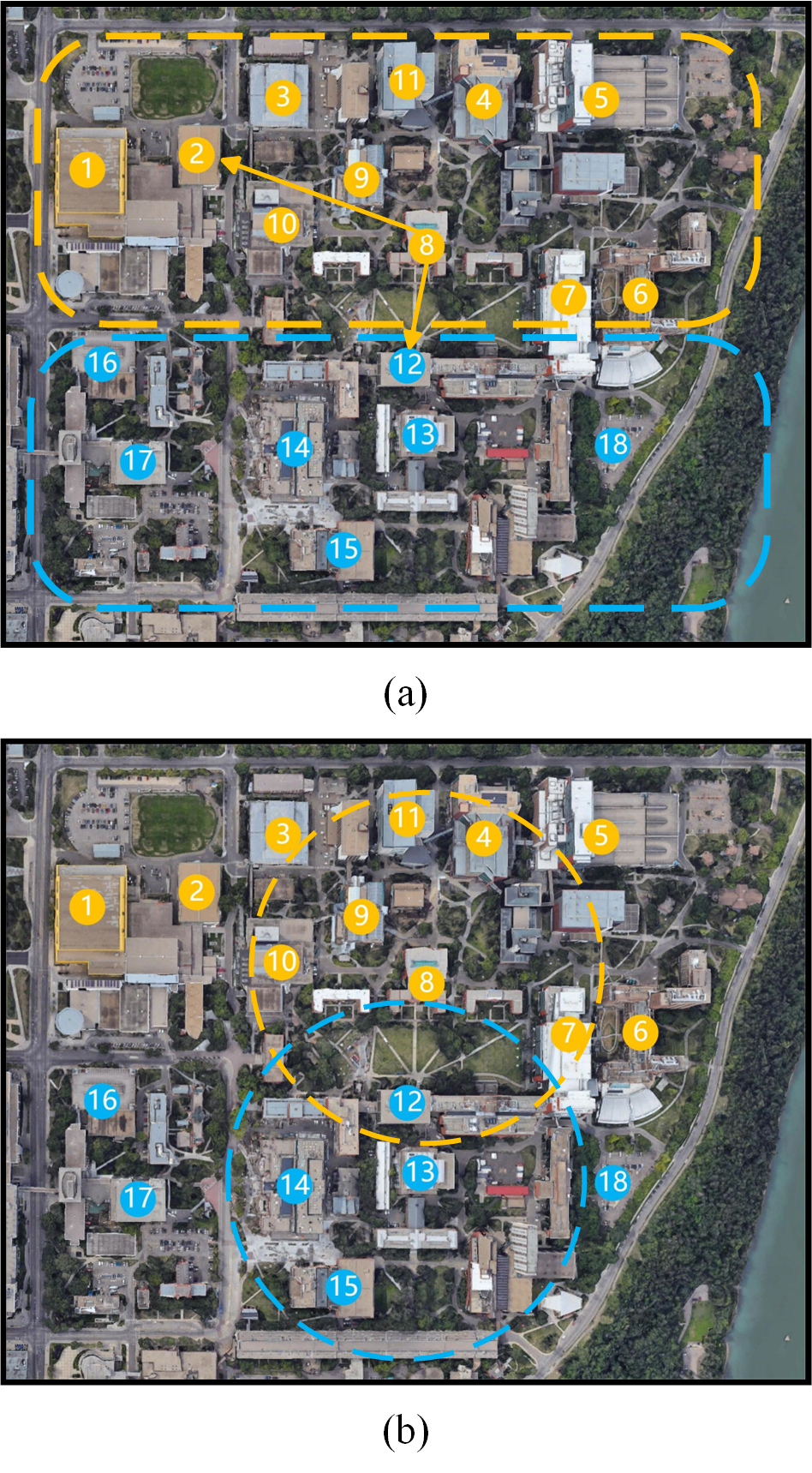}
  \caption{Spatial annotation schemes.
  (a) Conventional annotation scheme in HR. 
  The area is divided into non-overlapping subregions according to geographic location. 
  Buildings within each region are assigned multi-scale labels according to membership. 
  A hierarchical semantic label tree is constructed to represent correspondences across scales.
  (b) Distance-based spatial annotation scheme. 
  Each building serves as the center, and surrounding buildings are ranked by geographic distance to form the ground truth for retrieval. 
  Different distance thresholds (circles) are adopted to define multiple relevance levels for hierarchical retrieval evaluation.
  }
  \label{fig_2}
  \end{figure}

\subsection{Metric Learning}
Metric learning underpins cross-view geo-localization by constructing an embedding space where images from different domains can be compared meaningfully, thereby enhancing generalization beyond the training set~\cite{7298682}. 
To achieve this, a variety of loss functions, such as CosFace~\cite{8578650}, ArcFace~\cite{9449988}, Triplet Loss~\cite{7298682,9068282}, Contrastive Loss~\cite{1640964}, N-pair Loss~\cite{NIPS2016_6b180037} and Instance Loss~\cite{8237667,10.1145/3383184}, have been developed to encourage a geometrically well-structured feature space.
These approaches have been widely adopted in various tasks such as face recognition, person re-identification, and vehicle re-identification, contributing to substantial improvements in retrieval accuracy.

Recently, DyML has emerged as a promising direction of retrieval tasks, aiming to learn a unified metric space that adapts to multiple semantic scales.
Differing from hierarchical classification~\cite{10.5555/3495724.3495862,7780493}, DyML is formulated in an open-set regime, where the set of classes during test is disjoint from those during training.
Sun et al.~\cite{9578883} formally introduced the DyML problem and established three multi-scale retrieval datasets as standard evaluation benchmarks.
They also proposed the Cross-Scale Learning (CSL) framework as a strong baseline in this task.
The HAPPIER approach~\cite{10.1007/978-3-031-19781-9_15}, which currently represents the state of the art in DyML, directly optimizes hierarchical average precision by employing differentiable surrogate functions.
Note that although these methods excel at multi-scale retrieval, they are not inherently tailored for cross-view geo-localization. To bridge this gap, we adapt dynamic metric learning to the cross-view setting by introducing a novel DyCL framework thereby enhancing HR performance in cross-view geo-localization.

\section{Distance-Aware Campus Dataset}
\subsection{Data Collection}
Existing benchmarks for cross-view geo-localization are limited by their reliance on simple semantic labels, which fail to capture hierarchical relationships or spatial continuity between locations.
From the perspective of DACVGL, it is desirable that not only the image annotations reflect relative spatial relationships, but also that the image content can provide rich contextual information beyond the primary target.
Compared to ground-level imagery, drone imagery is better suited to these requirements, as it naturally includes a broader background and more comprehensive environmental details.
An existing dataset which incorporates drone imagery is University-1652~\cite{zheng2020university,zheng2023uavm}, however it lacks hierarchical and distance-based annotations.
To address these limitations, we establish the DA-Campus dataset, featuring multi-source, multi-view imagery with explicit hierarchical and distance-aware labels.

Following the similar data collection procedures of University-1652, we constructed a large-scale, spatially comprehensive dataset for cross-view geo-localization, named Distance-Aware Campus (DA-Campus). 
The dataset consists of satellite-view images collected from Google Maps and synthetic drone-view images sampled from Google Earth 3D models.
To establish cross-source correspondences, we first collected metadata for university buildings from Wikipedia, including building names and their respective university affiliations.
These building names were then geocoded into precise geographic coordinates (latitude and longitude).
In total, 750 buildings were selected, with 450 designated for training and 300 for testing.
The geographic coordinates were exploited to extract corresponding satellite-view images from Google Maps.
To acquire drone-view imagery without incurring the cost of real-world drone flights, we leveraged 3D models from Google Earth to simulate drone footage.
For each building, a virtual drone was controlled to follow a 360-degree circular path, resulting in 60 images captured from evenly spaced viewpoints around the target.

\subsection{Spatial Annotation Scheme}
To precisely annotate the relative spatial relationships among the collected building images, we considered two alternative schemes, as illustrated in Fig.~\ref{fig_2}.
The first scheme, as shown in Fig.~\ref{fig_2}(a) follows the standard approach commonly used in dynamic metric learning tasks, where buildings are grouped into discrete regions.
A hierarchical label tree is constructed to represent the semantic correspondence between different scales.
Despite this method facilitates clear semantic organization, there is a notable problem arising with buildings located near region boundaries. 
These buildings often share similar visual features to those in adjacent regions. 
Nevertheless, they will be labeled as negative samples solely due to region assignments.
For instance, Building 8 and Building 12 in Fig.~\ref{fig_2} are geographically close and share many similar visual features from their surrounding environment.
However, according to semantic correspondence, they should be assigned to different regions.
In contrast, more distant buildings, such as Building 2 and Building 8, may be grouped into the same region.
Such inconsistencies frequently arise when spatial partitions are adjacent.

Due to this geographical spatial continuity problem, we finally adopted the distance-based spatial annotation scheme illustrated in Fig.~2(b).
Although this method requires calculating the spatial relevance between every pair of buildings, it is better suited for capturing true geographic relationships and supporting the study of DACVGL.
Specifically, we manually annotated each building with GPS coordinates and computed the Euclidean distance between all building pairs.
For each building, a ranking list of all other buildings was constructed according to geographic distance.
The ranking list serves as the ground truth for evaluating retrieval performance in the subsequent tasks.
\begin{table}[!t]
\caption{Detailed features of DA-Campus dataset.}                       
\centering  
   \label{tab_1}
  \renewcommand{\arraystretch}{1.5}
\begin{tabular}{c|cc}                        
\toprule                                     
\multirow{2}{*}{Dataset} & \multicolumn{2}{c}{DA-Campus} \\   
\cmidrule{2-3}   
& Training Set & Test Set \\
\midrule 
Buildings & 450 & 300 \\   
Images &  27.4k & 18.3k \\
\midrule 
Platform/Viewpoint & \multicolumn{2}{c}{Drone, Satellite} \\    
GPS & \multicolumn{2}{c}{$\checkmark$} \\         
Evaluation Protocol & \multicolumn{2}{c}{R@K, mAP, H-AP, ASI, NDCG} \\ 
\bottomrule                                 
\end{tabular}
\end{table}

\subsection{Evaluation Protocol}
From the perspective of cross-view geo-localization, DA-Campus supports two standard tasks: drone navigation (Satellite$\rightarrow$Drone) and drone localization (Drone$\rightarrow$Satellite).
Leveraging the available distance information, both tasks enable evaluation under a HR protocol.
Specifically, for each query, buildings in the ranking list are categorized into different relevance levels based on distance thresholds of 0 m, 200 m, and 500 m, corresponding to small, middle, and large retrieval scales.
The detailed formulation is described in Section~\ref{sec_tf}.
At each scale, conventional metrics such as Recall@K (R@K) and mean Average Precision (mAP) can be computed.
In addition, three hierarchical metrics, Hierarchical Average Precision (H-AP)~\cite{10.1007/978-3-031-19781-9_15}, Average Set Intersection (ASI)~\cite{9578883}, and Normalized Discounted Cumulative Gain (NDCG)~\cite{10.1145/582415.582418}, are implemented to provide a comprehensive analysis of retrieval performance across spatial scales.
The main characteristics of the DA-Campus dataset are summarized in Table~\ref{tab_1}.

\section{Distance-Aware Cross-View Geo-Localization}
\subsection{Task Formulation}
\label{sec_tf}
\begin{figure}[!t]
  \centering
  \includegraphics[width=3.2in]{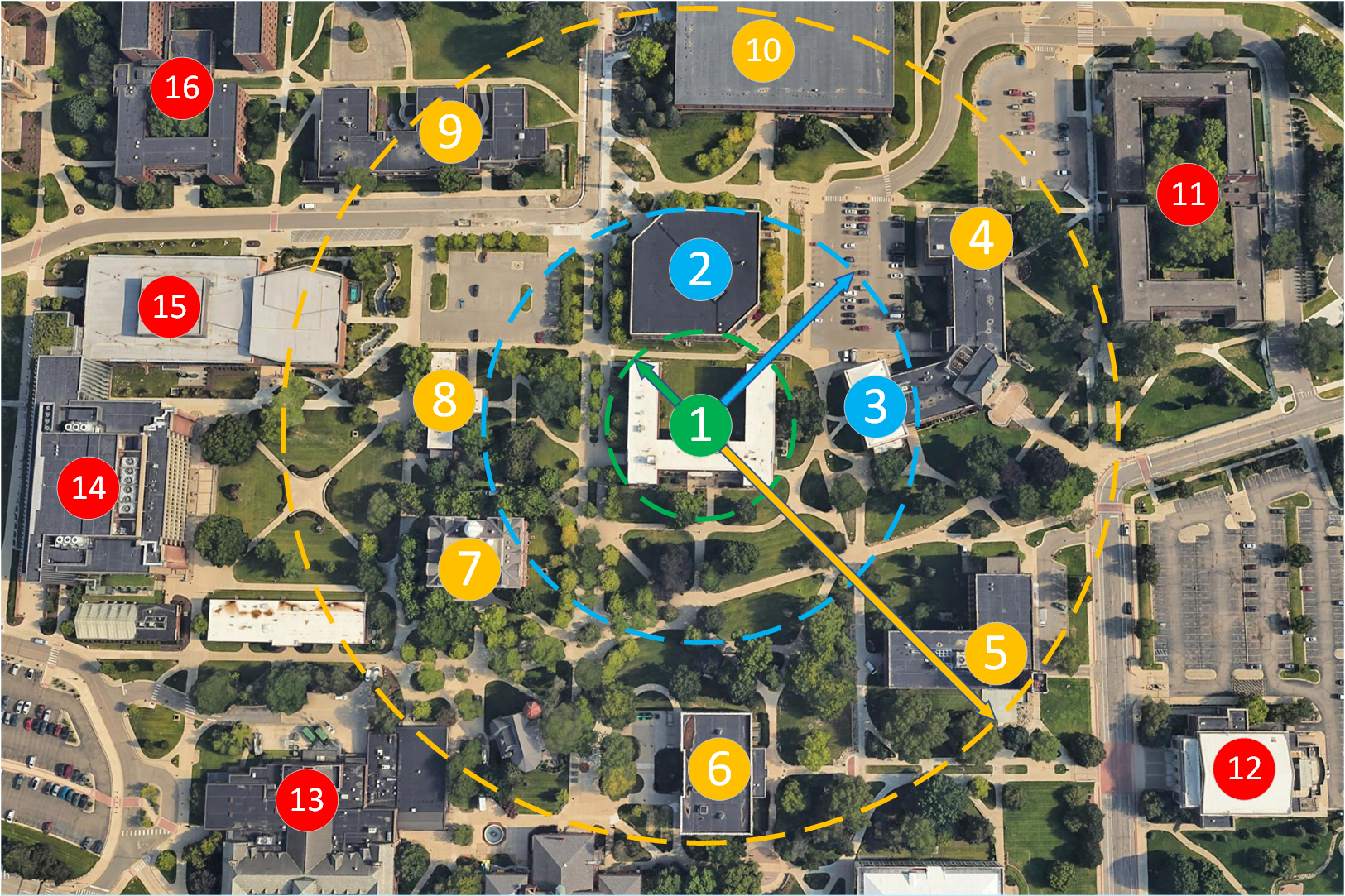}
  \caption{Illustration of the detailed spatial partitioning strategy in DACVGL.
  Buildings surrounding a reference building (Building 1) are partitioned into different geographic scales according to distance thresholds, as indicated by concentric colored circles.
  The green circle represents the reference building itself (distance = 0), while blue, yellow, and red circles denote neighboring buildings at increasing distance ranges respectively. 
  }
  \label{fig_3}
  \end{figure}
We formulate the DACVGL task in a manner analogous to HR as follows.
Let us assume a retrieval set $\mit\Omega^A=\{I_1^A,I_2^A,\ldots,I_C^A\}$ comprising images from $C$ buildings.
Each $I_c^A$ denotes the set of images of the $c$-th building captured from viewpoint A.
As illustrated in Fig.~\ref{fig_3}, for a randomly selected building, e.g., Building 1, the set of its own images forms the smallest geographic scale, denoted as $\mathcal{S}_c^0$.
Then its nearest neighbors, such as Buildings 2 and 3, are considered to form the geographic scale $\mathcal{S}_c^1$.
The distances between Building 1 and these buildings are less than the smallest positive threshold. 
By comparison, buildings that are farther away, such as Buildings 4–10, form the larger geographic scale $\mathcal{S}_c^2$, and so on.
By applying a set of increasing distance thresholds, we define a series of nested geographic ranges $\mathbb{S}_c=\{\mathcal{S}_c^l|l\in[0,L]\}$.
Each $\mathcal{S}_c^l$ contains a total of $N_c^l$ labeled images of buildings from a subset $C_c^l$ of $C$, i.e.,
\begin{equation}
  \label{eq_0}
  \begin{aligned}
    \mathcal{S}_c^l=\{(x_j^A,y_j)|j=1,2,\ldots,N_i^l,y_j\in{C}_c^l\},
  \end{aligned}
\end{equation}
where $x_j^A$ indicates that images are captured from viewpoint A.

Given a query image $q_i^B$ of the $c$-th building captured from viewpoint $B$, the goal of DACVGL is to learn a unified feature space $\mathcal{F}^{A\leftrightarrow{B}}$ in which cross-view images can be directly compared across all geographic scales in $\mathbb{S}_c$.
Specifically, the similarity between $q_i^B$ and candidate image $x_j^A$ can be computed in $\mathcal{F}^{A\leftrightarrow{B}}$ and denoted as $r_{ij} = \text{sim}(q_i^B, x_j^A)$.
At each scale $\mathcal{S}_c^l$, we define two sets: $\mathcal{S}_c^{\leq{l}}=\bigcup_{m=0}^{l}\mathcal{S}_c^m$, and its complement $\mathcal{S}_c^{>l}=\mathbb{S}_c \setminus \mathcal{S}_c^{\leq{l}}$.
Formally, the objective function here can be formulated as:
\begin{equation}
\label{eq_1}
\begin{aligned}
\max(r_{ij} - r_{ik}),\ \  \forall\, \mathcal{S}_c^l,\; (x_j^A,y_j) \in \mathcal{S}_c^{\leq{l}},\; (x_k^A,y_k) \in \mathcal{S}_c^{>l}.
\end{aligned}
\end{equation}
This learning objective ensures that, at each geographic scale, images more relevant to the query are assigned higher similarity scores, and vice versa.
As a result, the model is encouraged to capture the hierarchical structure of spatial relevance in cross-view retrieval. 
In the final ranking, images of the same building or nearby buildings are consistently prioritized over those that are more distant, yielding a distance-aware retrieval paradigm.

We choose a threshold-based partitioning strategy instead of direct distance-based ranking for two main reasons.
First, in practical scenarios, it is often meaningless to impose a strict ordering of images by precise distance, especially when buildings are distributed in different directions.
Grouping buildings by distance thresholds better reflects local visual context and feature correlations.
Second, this threshold-based partitioning aligns with established evaluation protocols in HR, thereby facilitating a more rigorous and interpretable assessment of retrieval performance across multiple scales.
Moreover, it is worth noting that the proposed anchor-specific definition of scales $\mathbb{S}_c$ highlights an essential distinction between DACVGL and standard HR: in DACVGL, visual and spatial correlations are always defined dynamically and continuously over each reference building.
It is impossible to organize all images into a fixed, mutually exclusive semantic hierarchy as in standard HR.

\subsection{Baseline Approach}
\begin{figure*}[!t]
  \centering
  \includegraphics[width=0.95\textwidth]{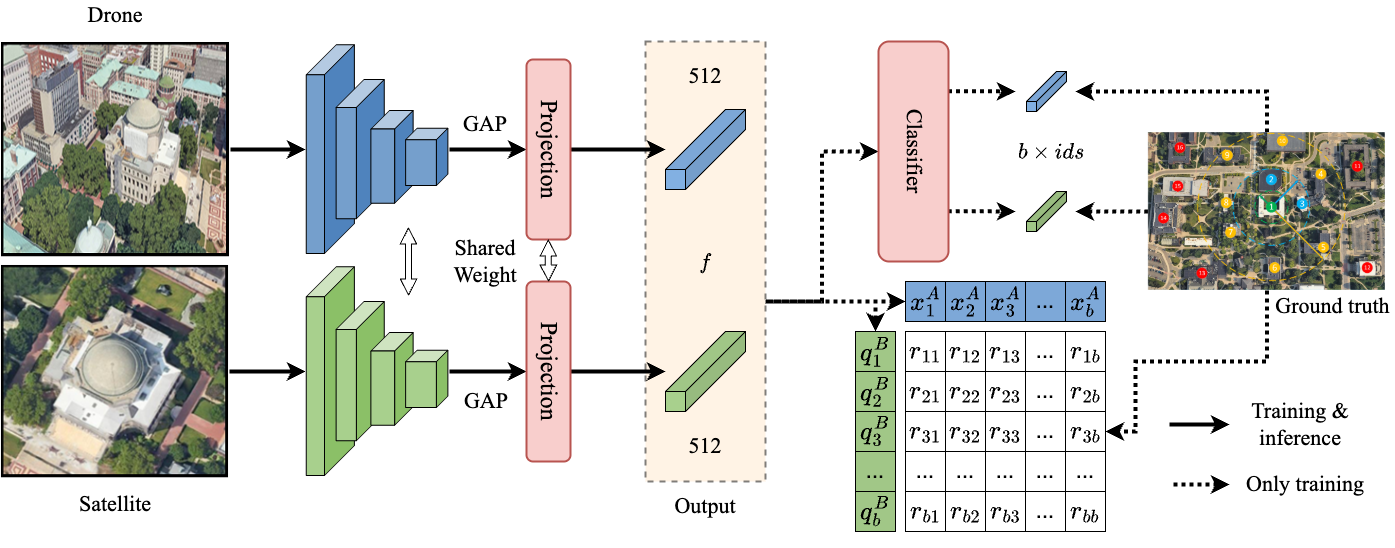}
  \caption{Overview of the baseline cross-view geo-localization framework. 
  Images from drone and satellite views are processed by a shared-weight backbone, followed by global average pooling (GAP) and a shared projection layer to generate 512-dimensional feature embeddings.
  The embeddings are used both for retrieval, via pairwise similarity $r_{ij}$, and for supervised training with an additional classifier and supervision from spatial annotations. 
  Solid arrows denote data flow during both training and inference, while dashed arrows mean modules are only valid in training. 
  Finally, the embedded feature $f$ are treated as output of the whole model for retrieval, with classification branch discarded.}
  \label{fig:baseline}
  \end{figure*}
As illustrated in Fig.~\ref{fig:baseline}, we adopt a baseline framework analogous to standard cross-view retrieval method to learn a distance-aware retrieval model. 
Given input images ${x_i}$ captured from different viewpoints, a siamese backbone network is exploited to extract feature representations ${f_i}$ for each image. 
The backbone can be instantiated with various pre-trained models, with the original classification head removed. 
After global average pooling (GAP), a fully connected projection layer is appended to map all features into a embedding space with unified dimensionality.
Differing from conventional cross-view retrieval frameworks, our approach incorporates hierarchical supervision during training. 
Specifically, each batch consists of pairs of images $(x_i^A, x_i^B)$, where each pair is captured from same building with different views. 
For the smallest scale, i.e., $\mathcal{S}_i^0$, the extracted features ${f_i}$ are further passed through an additional classifier. 
The associated building labels ${y_i}$ are used as supervision to train this classification branch.
For other scales, each sample in the batch serves as an anchor.
Contrastive losses are applied on similarities between samples at each scale to comprehensively capture hierarchical spatial relationships.

\subsection{Dynamic Contrastive Learning}
\label{sec:dycl}
\begin{figure*}[!t]
  \centering
  \includegraphics[width=0.95\textwidth]{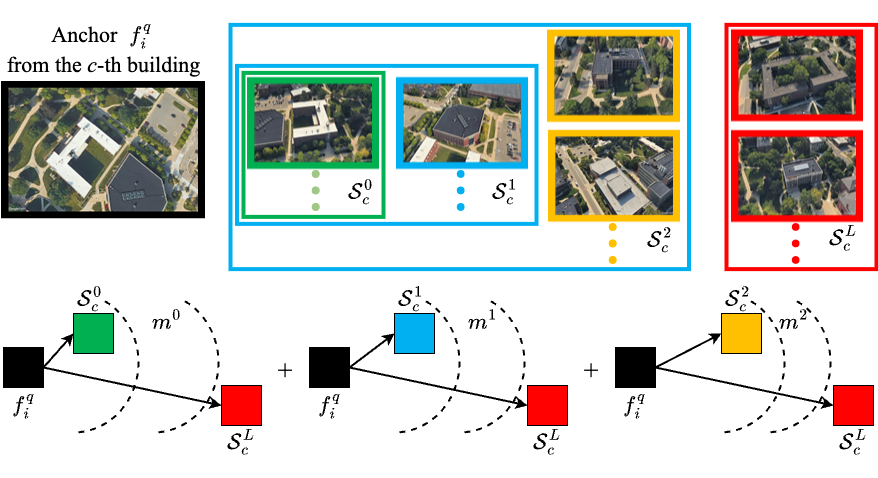}
  \caption{Illustration of the DyCL margin control mechanism across multiple geographic scales. 
  Given an anchor sample $f_i^q$, the green, blue, and yellow squares represent reference samples at increasing distances, corresponding to different geographic scales. 
  The red squares denote pure negative samples, i.e., images of buildings whose distances exceed the largest threshold.
  During learning at each scale, DyCL explicitly controls the similarities between the anchor (black square) and the reference samples (colored squares).
  Here, a shorter distance between two squares indicates higher similarity.
  It can be seen that the similarity between $f_i^q$ and $\mathcal{S}_c^L$ remains fixed as a stable reference, while the similarities to other samples decrease gradually with increasing scale as determined by the corresponding $m^l$s.
}
  \label{fig:dycl}
  \end{figure*}
A typical challenge in HR is the inherent conflict that arises when metric learning is performed across different scales within a unified feature space.
As can be seen from Eq.~(\ref{eq_1}), without any extra constraints, optimizing $\max(r_{ij} - r_{ik})$ at scale $\mathcal{S}_c^l$ causes the samples in $\mathcal{S}_c^l$ to be included in $r_{ij}$. 
However, at scale $\mathcal{S}_c^{l-1}$, the same samples are included in $r_{ik}$, resulting in directly opposing optimization objectives at adjacent scales.
Our experiments in Section~\ref{sec:single_vs_multi} also empirically verify this observation.
To address this issue in the DACVGL scenario, we propose a Dynamic Contrastive Learning (DyCL) loss function. 
DyCL directly optimizes the similarity among the anchor, positive, and pure negative samples at each scale. 
By explicitly controlling the similarity margins for different scales as a series of decreasing values, we mitigate the contradictions in Eq.~(\ref{eq_1}) and enhance the generalization ability of the model across multiple scales.

Specifically, for a given query image with output feature vector $f_i^q$ as the anchor, we employ cosine similarity, $r_{ij} = {f_i^q}^Tf_j$, as the metric between different samples. 
At each scale $\mathcal{S}_c^l$, the similarity between the anchor and positive samples, namely $(f_j,y_j) \in \mathcal{S}c^{\leq{l}}$, is denoted as $r_{p,j}^l$.
The similarity between the anchor and pure negative samples is denoted as $r_{n,k}^L$.
Here pure negatives $(f_k, y_k) \in \mathcal{S}_c^L$ correspond to images of buildings whose geographic distances to the anchor exceed the largest threshold.
DyCL seeks to enforce a margin between $r_{p,j}^l$ and $r_{n,k}^L$ as: 
\begin{equation}
  \label{eq_rpjnk}
  \begin{aligned}
    r_{p,j}^l-r_{n,k}^L\geq{m^l},\ \ l=0,1,\ldots,L-1,
  \end{aligned}
\end{equation}
where $m^l$s are the margins associated with each scale.
Intuitively, we set $m^0>m^1>\cdots>m^{L-1}>0$ to explicitly reflect the variation in geographic distance across scales.
Therefore, the DyCL loss can be formally defined as:
\begin{equation}
  \label{eq_dycl}
  \begin{aligned}
    \mathcal{L}_{\mathrm{DyCL}} = \sum_{l=0}^{L-1} \log (1 + \sum_{j=1}^{|\mathcal{S}_c^{\leq l}|} \sum_{k=1}^{|\mathcal{S}_c^L|} \exp\tau (r_{n,k}^L - r_{p,j}^l + m^l) ),
  \end{aligned}
\end{equation}
where $\tau$ is a scaling factor and $|\cdot|$ denotes the cardinality of a set.
Finally, the entire loss function follows a symmetric cross-view retrieval paradigm: images from either viewpoint can be used as anchors, with their cross-view counterparts in $\mathbb{S}_c$ serving as reference samples for loss computation.
This symmetric formulation allows the loss to be applied in both directions, ensuring comprehensive cross-view training.  
Since each batch is constructed by pairing images from different views, all required triplets for Eq.~(\ref{eq_dycl}) can be efficiently generated within a single batch.

DyCL takes inspiration from the idea Cross-Scale Learning (CSL)~\cite{9578883}, yet with explicit difference.
CSL adopts a typical metric learning approach similar to CosFace~\cite{8578650}, where the classification branch serves as a set of proxies for the class centers during training.
The similarity is calculated as the inner product between the output feature and the classifier weights.
In contrast, DyCL adopts an explicit contrastive learning approach, directly comparing different samples within a batch. 
We argue that such query-centered direct comparison is crucial for anchor-specific tasks like DACVGL.
On the other hand, compared with HAPPIER~\cite{10.1007/978-3-031-19781-9_15}, another multi-scale contrastive loss, DyCL chooses to explicitly control the margins between samples at different scales, rather than optimizing the ranking metric as in HAPPIER. 
Our comparisons in Section~\ref{sec:da_campus} demonstrate that DyCL and HAPPIER exhibit strong complementarity. 
Moreover, their combination yields further performance gains.


\textbf{Clustering loss.}
At the scale $\mathcal{S}_c^0$, we additionally employ a loss to ensure that instances from the same building are closely clustered:
\begin{equation}
  \label{eq_clust}
  \begin{aligned}
    \mathcal{L}_{\text{clust}} = -\log\left(\frac{\exp(w_i^T f_i)}{\sum_{j=1}^C \exp(w_j^T f_i)}\right),
  \end{aligned}
\end{equation}
where $w_i$ is the normalized proxy corresponding to the fine-grained class of feature $f_i$.
This normalized variant of cross-entropy loss~\cite{zhai2019classificationstrongbaselinedeep,10.1007/978-3-030-58586-0_27} essentially enforces the clustering of instances across the entire training set according to their labels.
It compensates for the limitation that in DyCL all sample comparisons are restricted within a single batch.

\subsection{Multl-Scale Re-ranking}
Re-ranking is a prevailing post-processing step in image retrieval, designed to refine the initial ranking by leveraging the structural relationships within the retrieval set. 
In the literature, discussions of re-ranking algorithms are normally based on a distance metric $d(q^B, x_j^A)$ defined on the learned feature embedding. 
To maintain consistency, we adopt the same notation in this section, instead of the inversely related feature similarity used in previous discussions.
Given the original distance $d(q^B, x_j^A)$, standard re-ranking algorithm fuses it with a smoothed Jaccard distance, resulting in the re-ranked distance $d_s^{\star}(q^B, x_j^A; k)$~\cite{8099872}.
Here $k$ is hyper-parameter controlling the size of reciprocal neighborhood set.

Existing re-ranking approaches typically fix $k=20$ to match the expected number of positive samples in standard benchmarks. 
This is because a proper neighborhood size $k$ should cover the main distribution of positive samples, so as to ensure that the reciprocal sets used for Jaccard distance computation are more discriminative.
However, in multi-scale retrieval, the number of relevant samples increases significantly at larger scales, making a fixed $k$ suboptimal.
Our experiments also reveal that re-ranking is highly sensitive to the choice of $k$ at different scales.
To address this, we propose a Multi-Scale Re-ranking (MSRerank) scheme, which repeatedly applies the standard re-ranking module with different $k$ parameters.
The $k$ values are selected based on prior knowledge of the building distribution in the training set as:
\begin{equation}
  \label{eq_k1}
  \begin{aligned}
    k^{l} = \max \left( 20,\, \frac{\mu}{C} \sum_{c=1}^C \left| \mathcal{S}_c^{\leq l} \right| \right),
  \end{aligned}
\end{equation}
where $\mu$ is an empirical hyper-parameter. In our experiments, we simply fix $\mu=0.1$.
Building on these re-ranking results, we introduce a segmented accumulative re-ranking algorithm to generate the final distance as described in Algorithm~\ref{alg:rerank}. 
\begin{algorithm}[htbp]
  \caption{Multl-Scale Re-ranking Algorithm.}
  \label{alg:rerank}
  \begin{algorithmic}[1]
  \STATE \textbf{Input:} Distance matrix $D\in\mathbb{R}^{(|\mit\Omega^A|+1)\times(|\mit\Omega^A|+1)}$ between $q^B$ and each entry in $\mit\Omega^A$, parameter list $\{k^l\}_{l=0}^{L-1}$.
  \STATE \textbf{Output:} Final re-ranked distances $d^{\star}(q^B,x_j^A)$ for $x_j^A\in\mit\Omega^A$.
  
  \STATE Initialise $d^{\star}(q^B,x_j^A)\leftarrow 0$, Mask $\mathcal{M}\leftarrow\mit\Omega^A$.
  
  \FOR{$l=0$ \textbf{to} $L-1$}
      \STATE \textbf{(a)\;Standard re-ranking:}
      \[d_{l}^{\star}(q^B,x_j^A)\leftarrow{d}_s^{\star}(q^B, x_j^A; k^l).\]
  
      \STATE \textbf{(b)\;Update active entries:}
      \[\forall\,x_j^A\in\mathcal{M},\; d^{\star}(q^B,x_j^A)\; \mathrel{+}= d_{l}^{\star}(q^B,x_j^A).\]
  
      \STATE \textbf{(c)\;Select top-$k^{l}$ samples:} 
      \[
      \mathcal{S}^{l} \leftarrow \text{Top-}k^{l} \text{ samples in } \mit\Omega^A \text{ sorted by } d^{\star}(q^B,x_j^A).
      \]
  
      \STATE \textbf{(d)\;Mask update:} 
      \[
      \mathcal{M} \leftarrow \mathcal{M} \setminus \mathcal{S}^{l}.
      \]
  \ENDFOR
  \RETURN $d^{\star}(q^B,x_j^A).$
  \end{algorithmic}
  \end{algorithm}

The proposed MSRerank algorithm iteratively accumulates the results of standard re-ranking at each scale, and masks out the top-ranked samples after each step. 
This prevents the re-ranking operations at larger scales from interfering with the refined orderings already established at smaller scales.
To our best knowledge, MSRerank is the first re-ranking method tailored for multi-scale retrieval.
It and can be seamlessly plugged into various cross-domain and standard hierarchical retrieval frameworks. 
Our experiments demonstrate that it consistently yields performance gains across all evaluated settings.

\section{Experimental Results}
\subsection{Implementation Details}
We conducted extensive comparisons on the DA-Campus dataset to benchmark the performance of various methods for DACVGL. 
For each $\mathbb{S}_c$, we categorize reference buildings into four relevance levels based on their geographic distance to the anchor building, namely $L=3$.
Two primary tasks are considered: drone navigation (Satellite $\rightarrow$ Drone) and drone localization (Drone $\rightarrow$ Satellite).
Based on the siamese network structure shown in Fig.~\ref{fig:baseline}, we adopt ResNet-50~\cite{7780459,5206848} as the backbone and train the feature extraction model.
During training, a shared classifier is deployed to compute the clustering loss in Eq.~(\ref{eq_clust}).
The entire network is optimized using the Adam optimizer, with a batch size of 128 and a total of 20 epochs.
During training, all images are resized to $256 \times 256$ pixels. 
Basic data augmentation is applied, including horizontal flipping. 
For satellite view images, random rotations are also performed.
During testing, the trained CNN is used to extract features from different sources.
Ranking lists are generated from the features to compute the aforementioned metrics, such as H-AP, ASI, NDCG, as well as R@1 and mAP for each scale.

\subsection{Single-scale Learning vs. Multi-scale Learning} 
\label{sec:single_vs_multi}
\begin{figure}[!t]
  \centering
  \includegraphics[width=0.45\textwidth]{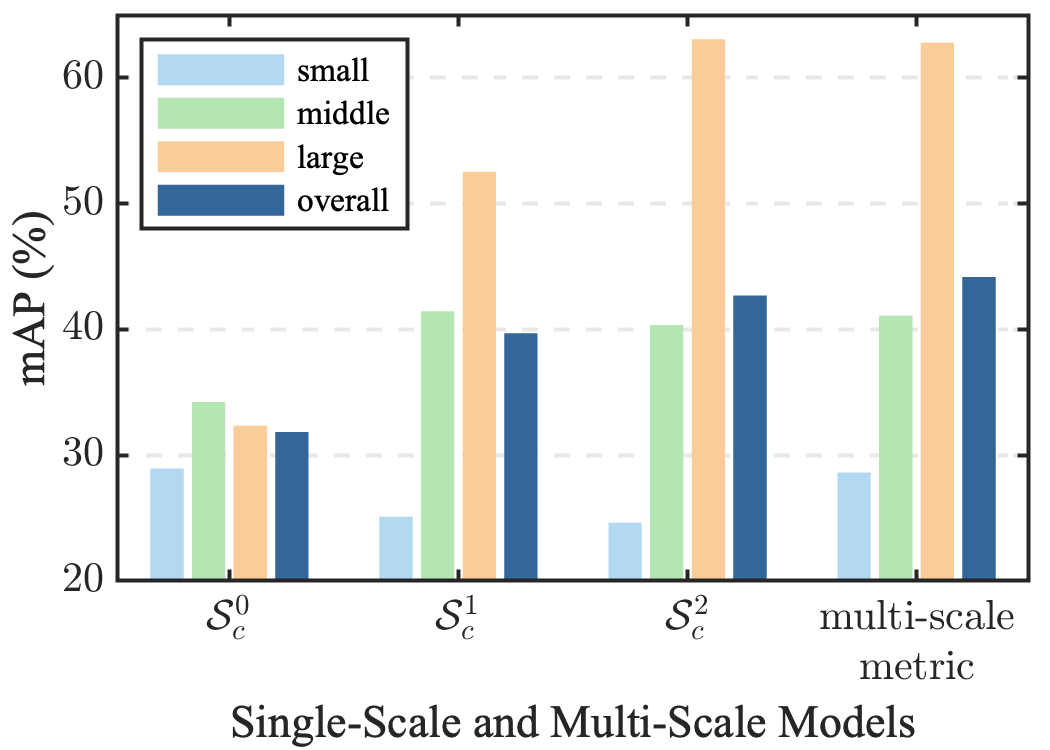}
  \caption{Comparison of performance among single-scale models and a multi-scale model on the drone navigation (Satellite $\rightarrow$ Drone) task.
  Each single-scale model is trained individually at scale $\mathcal{S}_c^l$, while the multi-scale model is jointly trained across all scales.
  For each model, mean Average Precision (mAP) is evaluated at the small, middle, and large scales on test set.
  We use the term \textit{Overall} to denote the average of three mAP values as the final comprehensive evaluation metric.
  }
  \label{fig:svsm}
  \end{figure}

Based on the DA-Campus dataset, we empirically validate the efficacy of multi-scale learning over single-scale learning. 
Specifically, for the drone navigation task, we first perform deep metric learning at each scale $\mathcal{S}_c^l$ separately and evaluate the resulting models across all scales. 
These single-scale models are then compared with a multi-scale learning model, which learns a unified embedding space encompassing all scales. 
For fair comparison, all models are trained exclusively using the conventional Triplet Loss~\cite{7298682,9068282}. 
The results are illustrated in Fig.~\ref{fig:svsm}. 
Here, we use terms small, middle, and large scales correspond to $S_c^0$, $S_c^1$, and $S_c^2$, respectively.
It can be observed that: 1) Each single-scale metric demonstrates relatively high accuracy within its specific scale. 
For example, the model trained only on $S_c^0$, achieves the highest performance in small scale testing. 
2) However, single-scale metric models do not generalize well to other spatial scales. 
For instance, the $S_c^0$ metric achieves only 32.34\% mAP on large scale testing, which is 30.74\% lower than the $S_c^2$ metric.
3) In contrast, the multi-scale metric learning model consistently outperforms the single-scale models, especially in terms of overall accuracy. It surpasses the $S_c^0$, $S_c^1$, and $S_c^2$ metrics by 12.28\%, 4.43\%, and 1.43\%, respectively, in the overall accuracy evaluation.
These results indicate that, even without employing targeted strategies such as DyCL, multi-scale metric learning is still necessary for improving overall retrieval accuracy.
      
\subsection{Main Results of DACVGL on DA-Campus}
\label{sec:da_campus}
We benchmark various methods on the DA-Campus dataset and present a comprehensive comparison with our proposed DyCL framework in Table~\ref{tab:main1}.
The candidate methods include both prevailing metric learning approaches~\cite{1640964,zhai2019classificationstrongbaselinedeep,7298682,9068282,10.1145/3383184,oord2019representationlearningcontrastivepredictive,pmlr-v139-radford21a,9578883,8954016,10.1007/978-3-031-19781-9_15} and state-of-the-art cross-view geo-localization methods~\cite{7299135,7780941,9360609,huangCVCities2024}.
For single-scale methods, models are trained on the small scale using building labels as supervision.
For CV-Cities~\cite{huangCVCities2024}, since official model weights are unavailable, we followed their released training pipeline and learned a model on DA-Campus with the pre-trained DINOv2~\cite{oquab2024dinov2learningrobustvisual} as backbone.
This approach yielded the strongest single-scale performance in our evaluation.
For the state-of-the-art hierarchical retrieval method HAPPIER~\cite{10.1007/978-3-031-19781-9_15}, we enhanced the original code by implementing the cross-view symmetric loss structure described in Section~\ref{sec:dycl}, and trained the corresponding model on DA-Campus.
Based on the same framework, we further combine the proposed DyCL with HAPPIER losses to learn a joint model, denoted as DyCL+HAPPIER in Table~\ref{tab:main1}.
\begin{table}[!t]
  \caption{Performance of different methods on three hierarchical evaluation metrics.
  For single-scale and multi-scale models, best and second-best results are highlighted in bold and underlined, respectively.
  Here, $\sum\text{Triplet}$ refers to the multi-scale metric learning described in Section~\ref{sec:single_vs_multi}, where triplet loss is applied at all scales directly.
  }  
  \label{tab:main1}
    \centering  
    \resizebox{\linewidth}{!}{  
    \renewcommand{\arraystretch}{1.5}
  \begin{tabular}{clccc|ccc}                         
  \toprule                                     
  \multicolumn{2}{c}{\multirow{2}*{Method}} & \multicolumn{3}{c|}{\textbf{Satellite$\rightarrow$Drone}} & \multicolumn{3}{c}{\textbf{Drone$\rightarrow$Satellite}}\cr     
  \cmidrule{3-5} \cmidrule{6-8}              
   & & H-AP & ASI & NDCG & H-AP & ASI & NDCG\\
  \midrule 
  \multirow{8}{*}{\rotatebox{90}{Single-Scale}} & Contrastive\cite{7299135,1640964} & 28.35 & 34.89 & 71.69 & 37.56 & 35.23 & 59.68 \\ 
   & NSM\cite{zhai2019classificationstrongbaselinedeep} & 30.88 & 37.56 & 72.90 & 39.23 & 37.07 & 60.16\\
   & Soft Margin Triplet\cite{7780941} & 30.72 & 41.32 & 73.36 & 41.82 & 38.96 & 62.03\\
   & Triplet\cite{7298682,9068282} & 31.14 & 38.02 & 73.21 & 40.82 & 38.73 & 61.32\\
   & Instance\cite{10.1145/3383184} & 38.23 & 47.01 & 76.71 & 45.86 & 46.18 & 64.30\\
   & LPN\cite{9360609} & 42.78 & 48.86 & 76.83 & 49.62 & 48.82 & 64.08\\
   & InfoNCE\cite{oord2019representationlearningcontrastivepredictive,pmlr-v139-radford21a} & \underline{43.37} & \underline{51.68} & \textbf{78.75} & \underline{50.38} & \underline{51.54} & \underline{67.38}\\
   & CV-Cities\cite{huangCVCities2024} & \textbf{45.00} & \textbf{54.78} & \underline{78.25} & \textbf{53.88} & \textbf{55.45} & \textbf{70.28}\\
  \midrule 
  \multirow{6}{*}{\rotatebox{90}{Multi-Scale}} & CSL\cite{9578883} & 13.12 & 19.01 & 60.24 & 16.56 & 14.44 & 38.23\\
   & Multi-Similarity\cite{8954016} & 36.32 & 44.52 & 71.2 & 46.79 & 40.91 & 63.64\\
   & $\sum\text{Triplet}$ & 44.82 & 49.43 & 77.45 & 50.97 & 44.98 & 66.28\\
   & HAPPIER\cite{10.1007/978-3-031-19781-9_15} & 44.95 & 53.75 & \underline {80.36} & 51.77 & 51.74 & \underline {69.34}\\
   & DyCL (ours) & \underline {46.02} & \underline {55.63} & 80.22 & \underline {52.74} & \underline {54.87} & 69.17\\
   & DyCL+HAPPIER (ours) & \textbf{47.73} & \textbf{56.27} & \textbf{81.06} & \textbf{54.35} & \textbf{55.95} & \textbf{70.63}\\
   \midrule 
   \multicolumn{2}{c}{DyCL+HAPPIER+MSRerank} & \textbf{49.97} & \textbf{59.64} & \textbf{81.26} & \textbf{56.32} & \textbf{57.16} & \textbf{71.06} \\
  \bottomrule                              
  \end{tabular} 
      }
  \end{table}

It can be observed that compared to the leading multi-scale method HAPPIER, DyCL achieves competitive results across most evaluation metrics. 
Most notably, the model jointly trained with DyCL and HAPPIER significantly outperforms all other methods.
Compared to the leading single-scale approach CV-Cities, it demonstrates an improvement of 2.73\% in H-AP, 1.49\% in ASI, and 2.81\% in NDCG for the Satellite $\rightarrow$ Drone task. 
For the Drone $\rightarrow$ Satellite task, the gains are 0.47\% in H-AP, 0.5\% in ASI, and 0.35\% in NDCG, respectively, even when using a more compact ResNet-50 backbone.

We also report the mAP and R@1 results of different methods across all relevance levels in Table~\ref{tab:main2}.
This comprehensive comparison enables a more detailed analysis of model performance and robustness at different spatial scales.
At the small scale, hierarchical methods achieve comparable mAP to CV-Cities, but CV-Cities exhibits superior R@1 performance.
This result is consistent with the design of single-scale methods, which specifically focus on fine-grained image matching accuracy.
However, multi-scale approaches consistently outperform single-scale baselines at larger scales, leading to higher overall accuracy.
This outcome aligns with the goal of pertinent cross-domain retrieval.
When fine-grained matching fails, it remains important to rank more geographically relevant images closer to the top, thus enhancing the practical utility of retrieval results.
Moreover, among leading multi-scale methods, both HAPPIER and DyCL deliver competitive results.
HAPPIER achieves the best mAPs and R@1s at the small scale, while DyCL demonstrates clear advantages at larger scales.
The model trained with the combination of DyCL and HAPPIER losses achieves improvements across all scales, indicating that the proposed DyCL is highly complementary to the HAPPIER in the scenario of hierarchical contrastive learning.
\begin{table*}[!t]
\caption{Performance of different methods in terms of mAP and R@1 across all spatial scales.
For both single-scale and multi-scale models, the best and second-best results are highlighted in bold and underlined, respectively.
Overall accuracies are calculated as the mean values of the results at the small, middle, and large scales.
} 
\label{tab:main2}
  \centering  
  \resizebox{\linewidth}{!}{  
  \renewcommand{\arraystretch}{1.5}
\begin{tabular}{clcc|cc|cc|cc||cc|cc|cc|cc}                        
\toprule                                    
\multicolumn{2}{c}{\multirow{3}*{Method}} & \multicolumn{8}{c||}{\textbf{Satellite$\rightarrow$Drone}} & \multicolumn{8}{c}{\textbf{Drone$\rightarrow$Satellite}}\cr     
\cmidrule{3-10}\cmidrule{11-18}             
 &  & \multicolumn{2}{c|}{Small} & \multicolumn{2}{c|}{Middle} & \multicolumn{2}{c|}{Large} & \multicolumn{2}{c||}{Overall} & \multicolumn{2}{c|}{Small} & \multicolumn{2}{c|}{Middle} & \multicolumn{2}{c|}{Large} & \multicolumn{2}{c}{Overall}\\
 &  & mAP & R@1 & mAP & R@1 & mAP & R@1 & mAP & R@1 & mAP & R@1 & mAP & R@1 & mAP & R@1 & mAP & R@1\\
\hline
\multirow{8}{*}{\rotatebox{90}{Single-Scale}} & Contrastive\cite{7299135,1640964} & 21.55 & 55.12 & 30.54 & 65.21 & 32.21 & 70.89 & 28.10 & 63.74 & 37.89 & 38.74 & 38.10 & 56.87 & 35.64 & 64.65 & 37.21 & 53.42\\	
 & NSM\cite{zhai2019classificationstrongbaselinedeep} & 31.45 & \underline{56.67} & 30.70 & 66.67 & 28.70 & 71.33 & 30.28 & 64.89 & 42.38 & \underline{47.56} & 38.22 & \underline{57.68} & 33.71 & 64.41 & 38.10 & 56.55\\
 & Soft Margin Triplet\cite{7780941} & 26.42 & 52.21 & 33.69 & 66.55 & 32.45 & 80.26 & 30.85 & 66.34 & 38.89 & 44.91 & 45.74 & 55.42 & 41.26 & 68.37 & 41.96 & 56.23\\
 & Triplet\cite{7298682,9068282} & 28.94 & 54.45 & 34.24 & 67.89 & 32.34 & 82.21 & 31.84 & 68.18 & 39.69 & 42.56 & 44.96 & 55.41 & 39.92 & 67.36 & 41.54 & 55.11\\
 & Instance\cite{10.1145/3383184} & 28.18 & 51.67 & 36.63 & 68.00 & 52.37 & 81.00 & 39.06 & 66.89 & 38.02 & 38.26 & 44.27 & 54.34 & 55.92 & 71.57 & 46.07 & 54.72\\
 & LPN\cite{9360609} & \underline{33.06} & 55.88 & 40.53 & \underline{72.56} & 54.39 & 82.62 & 42.66 & \underline{70.35} & \underline{44.58} & 43.20 & 47.36 & 58.22 & 56.24 & 72.21 & 49.39 & 57.88\\
 & InfoNCE\cite{oord2019representationlearningcontrastivepredictive,pmlr-v139-radford21a} & 30.91 & 54.33 & \underline{41.18} & 70.00 & \underline{59.51} & \underline{84.0} & \underline{43.87} & 69.44 & 40.53 & 41.23 & \underline{47.93} & 56.49 & \underline{63.28} & \textbf{76.96} & \underline{50.58} & \underline{58.23}\\
 & CV-Cities\cite{huangCVCities2024} & \textbf{34.71} & \textbf{67.81} & \textbf{43.18} & \textbf{77.71} & \textbf{59.80} & \textbf{84.70} & \textbf{45.89} & \textbf{76.74} & \textbf{45.53} & \textbf{53.53} & \textbf{51.65} & \textbf{67.65} & \textbf{65.19} & \underline{75.19} & \textbf{54.12} & \textbf{65.46}\\
\midrule 
\multirow{6}{*}{\rotatebox{90}{Multi-Scale}} & CSL\cite{9578883} & 13.66 & 39.33 & 12.74 & 45.00 & 12.34 & 49.67 & 13.04 & 44.67 & 17.77 & 17.77 & 15.87 & 21.57 & 14.36 & 25.41 & 16.00 & 21.58\\
& Multi-Similarity\cite{8954016} & 25.53 & 48.32 & 34.36 & 70.54 & 48.33 & 82.39 & 36.07 & 67.08 & 34.60 & 35.22 & 45.31 & 52.72 & 60.87 & 63.28 & 46.93 & 50.41\\
& $\sum\text{Triplet}$ & 28.63 & 62.69 & 41.10 & 72.73 & 62.79 & 84.21 & 44.17 & 73.21 & 38.36 & 46.00 & 49.65 & 62.31 & 65.63 & 80.21 & 51.21 & 62.84\\
& HAPPIER\cite{10.1007/978-3-031-19781-9_15} & \textbf{37.13} & \textbf{66.67} & 43.71 & \underline{76.33} & 54.66 & 85.67 & 45.17 & \underline{76.22} & \textbf{46.3} & \textbf{52.77} & 49.87 & \underline{66.04} & 58.68 & 79.78 & 51.62 & \underline{66.20}\\
& DyCL (ours) & 32.31 & 60.33 & \underline{44.56} & 76.00 & \textbf{64.79} & \underline{88.33} & \underline{47.22} & 74.89 & 41.36 & 44.03 & \underline{50.65} & 62.07 & \textbf{67.63} & \textbf{82.16} & \underline{53.21} & 62.75\\
& DyCL+HAPPIER (ours) & \underline{36.08} & \underline{64.67} & \textbf{45.77} & \textbf{78.33} & \underline{63.78} & \textbf{89.00} & \textbf{48.54} & \textbf{77.33} & \underline{44.87} & \underline{49.21} & \textbf{52.05} & \textbf{66.64} & \underline{66.89} & \underline{82.03} & \textbf{54.45} & \textbf{66.29}\\
\midrule 
\multicolumn{2}{c}{DyCL+HAPPIER+MSRerank}  & \textbf{39.33} & \textbf{67.08} & \textbf{45.96} & \textbf{78.64} & \textbf{64.86} & \textbf{89.92} & \textbf{50.05} & \textbf{78.81} & \textbf{47.32} & \textbf{53.11} & \textbf{54.09} & \textbf{67.71} & \textbf{67.91} & \textbf{83.21} & \textbf{56.44} & \textbf{67.37}\\
\bottomrule                                   
\end{tabular}
    }
\end{table*}

\subsection{Multi-scale Re-ranking}
\begin{figure}[!t]
  \centering
  \includegraphics[width=0.48\textwidth]{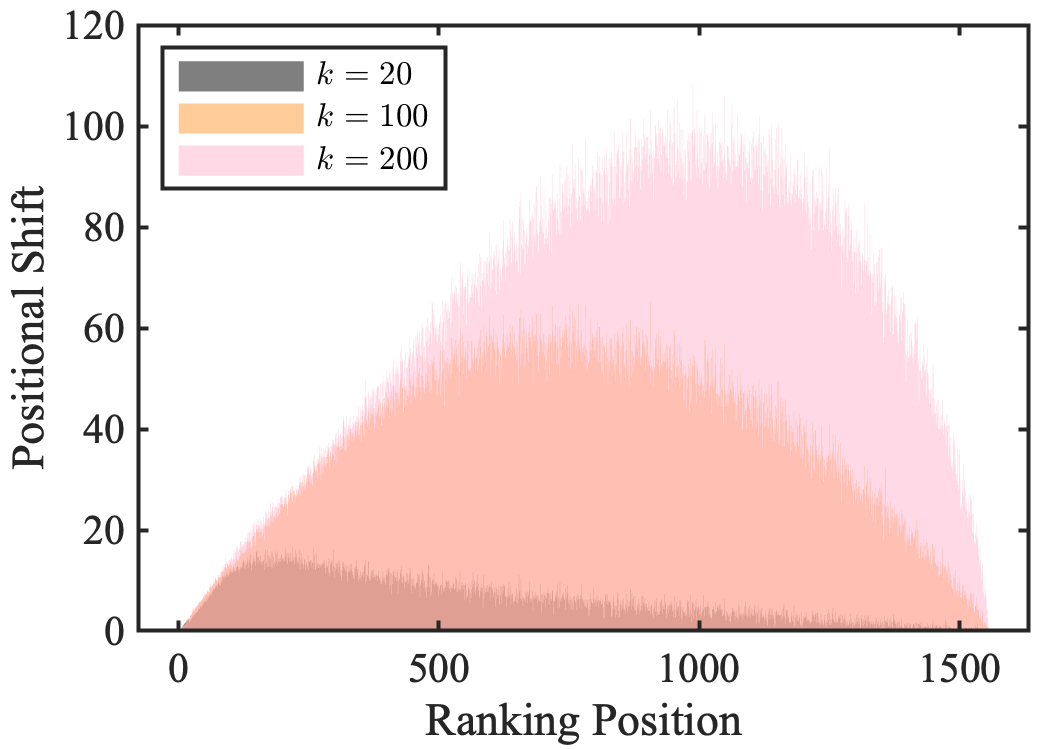}
  \caption{Distribution of position shifts across different ranking positions after re-ranking with various $k$s.
  On training set, every sample is treated as a query, ranking remaining samples as the retrieval set.
  Then the curves are obtained by averaging the position changes at each position.
  }
  \label{fig:rerank}
  \end{figure}
To quantitatively analyze the impact of different neighborhood sizes on re-ranking results, we performed re-ranking with different $k$s on the training set of DA-Campus.
Based on the Kendall Tau distance~\cite{kendall_distance} we then measured the positional shifts throughout the entire ranking list, as illustrated in Fig.~\ref{fig:rerank}.
The resulting plots intuitively show that varying $k$ influences different sections of the ranking list.
Also based on this observation, we empirically chose to select $k^l$s as described in Eq.~(\ref{eq_k1}) via prior knowledge from the training set.
Although $k^l$s are set according to training statistics, subsequent experiments demonstrate that the strategy generalizes well to test scenarios.
As shown in Tables~\ref{tab:main1} and~\ref{tab:main2}, the proposed MSRerank method delivers incremental but stable improvements across all evaluation metrics.
To further verify its efficacy and universality, we conducted more experiments on standard HR benchmark datasets, DyML-Vehicle, DyML-Animal and DyML-Product~\cite{9578883}, as summarized in Table~\ref{tab:rerank}.
It can be seen that MSRerank consistently improves the mAP over standard re-ranking results, while its R@1 performance remains equivalent due to the segmented accumulative strategy.
These comparisons demonstrate that MSRerank can serve as a generic post-processing technique to reliably enhance various hierarchical retrieval models.
It also aligns well with the core objective of hierarchical retrieval: to improve the quality of the entire ranking rather than focusing solely on the top-1 result.
\begin{table}[!t]
  \caption{Performance comparison between standard re-ranking and the proposed MSRerank algorithm on standard HR datasets.
  Here we adopt the trained HAPPIER as the baseline. 
  The best and second-best results are highlighted in bold and underlined, respectively.
}  
\label{tab:rerank}
\centering  
\resizebox{\linewidth}{!}{  
  \renewcommand{\arraystretch}{1.5}
  \begin{tabular}{lcc|cc|cc}                         
  \toprule                                     
  \multirow{2}*{Method} & \multicolumn{2}{c|}{\textbf{DyML-Vehicle}} & \multicolumn{2}{c|}{\textbf{DyML-Animal}} &\multicolumn{2}{c}{\textbf{DyML-Product}}\cr     
  \cmidrule{2-7}              
  & mAP & R@1 & mAP & R@1 & mAP & R@1\\
  \midrule 
  Triplet\cite{7298682,9068282} & 10.0 & 13.8 & 11.0 & 18.2 & 9.3 & 11.2 \\
  Multi-Similarity\cite{8954016} & 10.4 & 17.4 & 11.6 & 16.7 & 10.0 & 12.7 \\
  $\text{TL}_{\text{SH}}$\cite{8237571} & 26.1 & 84.0 & 37.5 & 66.3 & 36.32 & 69.6 \\
  NSM\cite{zhai2019classificationstrongbaselinedeep} & 27.7 & 88.7 & 38.8 & 69.6 & 35.6 & 57.4 \\
  $\sum\text{TL}_{\text{SH}}$\cite{8237571} & 25.5 & 81.0 & 38.9 & 65.9 & 36.9 & 58.5 \\
  $\sum\text{NSM}$\cite{zhai2019classificationstrongbaselinedeep} & 32.0 & \underline{89.4} & 42.6 & \textbf{70.0} & 36.8 & 60.8 \\
  CSL\cite{9578883} & 30.0 & 87.1 & 40 .8 & 60.9 & 31.1 & 52.7 \\
  HAPPIER\cite{10.1007/978-3-031-19781-9_15} & 37.0 & 89.1 & 43.8 & 68.9 & 38.0 & 63.7\\
  HAPPIER+Rerank\cite{8099872} & \underline{38.55} & \textbf{90.52} & \underline{44.53} & \underline{69.51} & \underline{41.21} & \textbf{64.5} \\
  HAPPIER+MSRerank & \textbf{40.7} & \textbf{90.52} & \textbf{46.44} & \underline{69.51} & \textbf{46.54} & \textbf{64.5} \\
  \bottomrule                              
  \end{tabular}
  } 
\end{table}

\subsection{Ablation Study of Hyper-parameters}
\label{sec:ablation}
\begin{figure}[!t]
  \centering
  \includegraphics[width=0.48\textwidth]{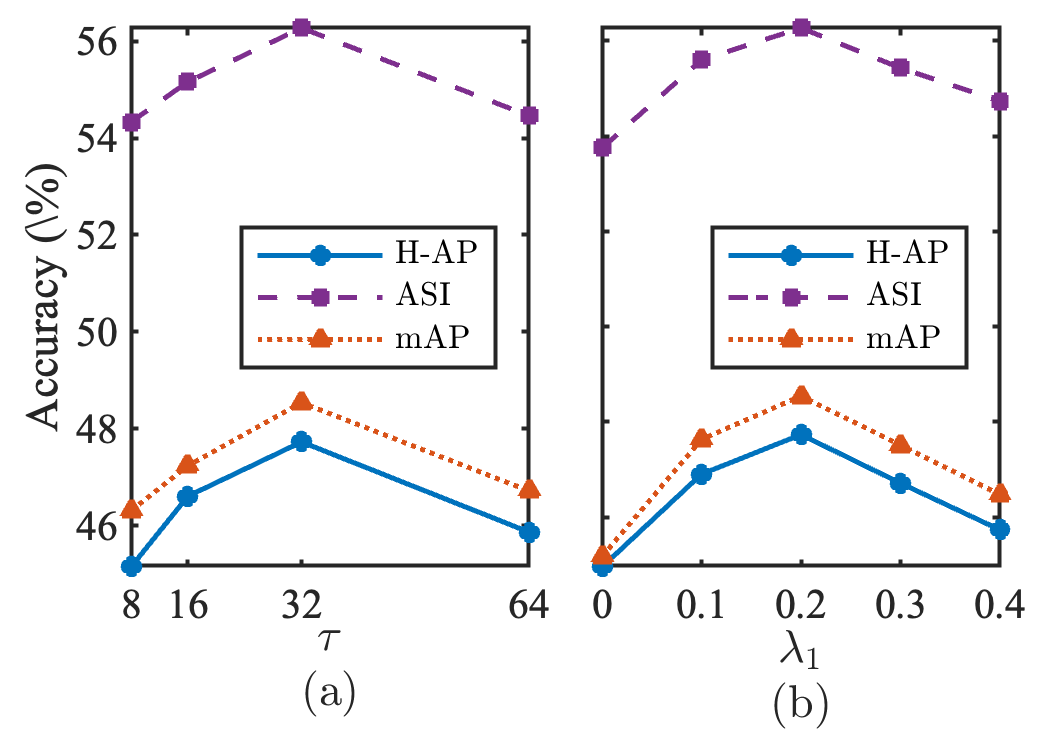} 
  \caption{Analysis of hyper-parameters $\tau$ and $\lambda$.
  For simplicity, only the results for the Satellite $\rightarrow$ Drone task are shown. 
  Similar trends are observed for the symmetric Drone $\rightarrow$ Satellite task.
  (a) Impact of scaling factor $\tau$ in Eq.~(\ref{eq_dycl}). 
  (b) Impact of loss weight $\lambda_1$ in Eq.~(\ref{eq_loss}).
  }
  \label{fig:params} 
  \end{figure}

We conduct ablation studies on key hyper-parameters of our framework, as illustrated in Fig.~\ref{fig:params}.
In Fig.\ref{fig:params} (a), we investigate the effect of the scaling factor $\tau$ in Eq.~(\ref{eq_dycl}).
We evaluate $\tau$ at values of 16, 32, and 64, and observe that the best overall performance is achieved at $\tau=32$.
We further analyze the loss composition of the best-performing DyCL+HAPPIER model.
The combined loss is defined as:
\begin{equation}
  \label{eq_loss}
  \begin{aligned}
    \mathcal{L}_{\text{total}} = \lambda_1\mathcal{L}_{\text{DyCL}} + \lambda_2\mathcal{L}_{\text{Clust}}+\lambda_3\mathcal{L}_{\text{HAPPIER}}.
  \end{aligned}
\end{equation}
We find that model performance is primarily affected by the proportion between $\mathcal{L}_{\text{DyCL}}$ and $\mathcal{L}_{\text{HAPPIER}}$, while $\mathcal{L}_{\text{Clust}}$ is relatively insensitive to weight variation.
Hence, we fixed $\lambda_2 = 0.1$ and $\lambda_3 = 0.9$, conducting an ablation study on $\lambda_1$ as in Fig.~\ref{fig:params} (b).
Base on the results, we empirically select $\lambda_1 = 0.2$, together with the above values of $\lambda_2$ and $\lambda_3$, as the optimal configuration for all of our experiments.

\section{Conclusion}
This paper revisits cross-view geo-localization from a distance-aware perspective and introduces the DACVGL task.
As a typical hierarchical contrastive learning problem, DACVGL exhibits two notable properties. 
First, supervision at multiple scales is inherently interrelated. 
Jointly learning embedding within a unified feature space benefits overall model performance.
Second, supervision contains complex structural information, which necessitates a contrastive learning approach rather than conventional metric learning during training.
To address these challenges, we construct the DA-Campus benchmark and propose a novel DyCL framework.
Experimental results demonstrate that DyCL yields consistent improvements in both hierarchical retrieval metrics and overall matching accuracy.
Beyond quantitative improvements, DyCL offers two practical advantages.
First, it helps make less severe mistakes, ensuring that top-ranked results are spatially closer to the target.
Second, if the model fails to retrieve the exact target at the top position, DyCL increases the likelihood that retrieved image contains potential clues about the target location.
Future works include further investigation of other hierarchical retrieval tasks in related domains that may similarly benefit from contrastive learning paradigms.
More validate of efficacy and flexibility of DyCL will be conducted in these scenarios.

\bibliographystyle{IEEEtran}
\bibliography{reference.bib}


 





\end{document}